\documentclass{article} 
\usepackage{iclr2021_conference,times}
\usepackage[normalem]{ulem}

\usepackage{amsmath,amsfonts,bm}

\def\eqref#1{equation~\ref{#1}}

\def\1{\bm{1}}

\DeclareMathAlphabet{\mathsfit}{\encodingdefault}{\sfdefault}{m}{sl}
\SetMathAlphabet{\mathsfit}{bold}{\encodingdefault}{\sfdefault}{bx}{n}

\usepackage{hyperref}
\usepackage{url}
\usepackage{eucal}
\usepackage{multicol}
\usepackage{tabularx}
\usepackage[titlenumbered, ruled,vlined]{algorithm2e}
\usepackage{times}
\usepackage{epsfig}
\usepackage{graphicx}
\usepackage{amsmath}
\usepackage{amssymb}
\usepackage{subcaption}
\usepackage{stfloats}
\usepackage{array}
\usepackage{booktabs}
\usepackage{color}
\usepackage{subcaption}
\usepackage[noend]{algpseudocode}
\usepackage{color, colortbl}
\usepackage{xspace, xcolor}
\usepackage{wrapfig}
\usepackage{subcaption}
\usepackage[capitalize]{cleveref}
\usepackage{wrapfig}
\usepackage{pgfplots}

\title{HyperSAGE: Generalizing Inductive \\Representation Learning on  Hypergraphs}

\author{Devanshu Arya, Deepak K. Gupta, Stevan Rudinac \& Marcel Worring 
\\
University of Amsterdam, Amsterdam, The Netherlands\\
\texttt{\{d.arya, d.k.gupta, s.rudinac, m.worring\}@uva.nl} \\
}

\begin{document}

\maketitle

\begin{abstract}
Graphs are the most ubiquitous form of structured data representation used in machine learning. They model, however, only pairwise relations between nodes and are not designed for encoding the higher-order relations found in many real-world datasets. To model such complex relations, hypergraphs have proven to be a natural representation. Learning the node representations in a hypergraph is more complex than in a graph as it involves information propagation at two levels: within every hyperedge and across the hyperedges. Most current approaches first transform a hypergraph structure to a graph for use in existing geometric deep learning algorithms. This transformation leads to information loss, and sub-optimal exploitation of the hypergraph's expressive power. We present HyperSAGE, a novel hypergraph learning framework that uses a two-level neural message passing strategy to accurately and efficiently propagate information through hypergraphs. The flexible design of HyperSAGE facilitates different ways of aggregating neighborhood information. Unlike the majority of related work which is transductive, our approach, inspired by the popular GraphSAGE method, is inductive. Thus, it can also be used on previously unseen nodes, facilitating deployment in problems such as evolving or partially observed hypergraphs. Through extensive experimentation, we show that HyperSAGE outperforms state-of-the-art hypergraph learning methods on representative benchmark datasets. We also demonstrate that the higher expressive power of HyperSAGE makes it more stable in learning node representations as compared to the alternatives. 



\end{abstract}

\section{Introduction}
Graphs are considered the most prevalent framework for discovering  useful information within a network, especially because of their capability to combine object-level information with the underlying inter-object relations \citep{wu2020comprehensive}. However, most structures encountered in practical applications form groups and relations that cannot be properly represented using pairwise connections alone, hence a graph may fail to capture the collective flow of information across objects. In addition, the underlying data structure might be evolving and only partially observed. Such dynamic higher-order relations occur in various domains, such as social networks \citep{tan2011using}, computational chemistry \citep{gu2020quantum}, neuroscience \citep{gu2017functional} and visual arts \citep{arya2019hyperlearn}, among others. These relations can be readily represented with \emph{hypergraphs}, where an edge can connect an arbitrary number of vertices as opposed to just two vertices in graphs. Hypergraphs thus provide a more flexible and natural framework to represent such multi-way relations \citep{wolf2016advantages}, however, this requires a representation learning technique that exploits the full expressive power of hypergraphs and can generalize on  unseen nodes from a partially observed hypergraph.

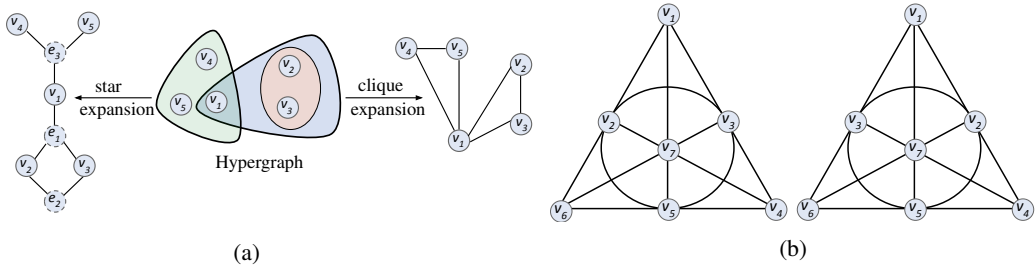
\begin{figure}
\begin{subfigure}{0.48\textwidth}
    \centering
    \begin{tikzpicture}
    \node () at (0,0) {\includegraphics[scale=0.28]{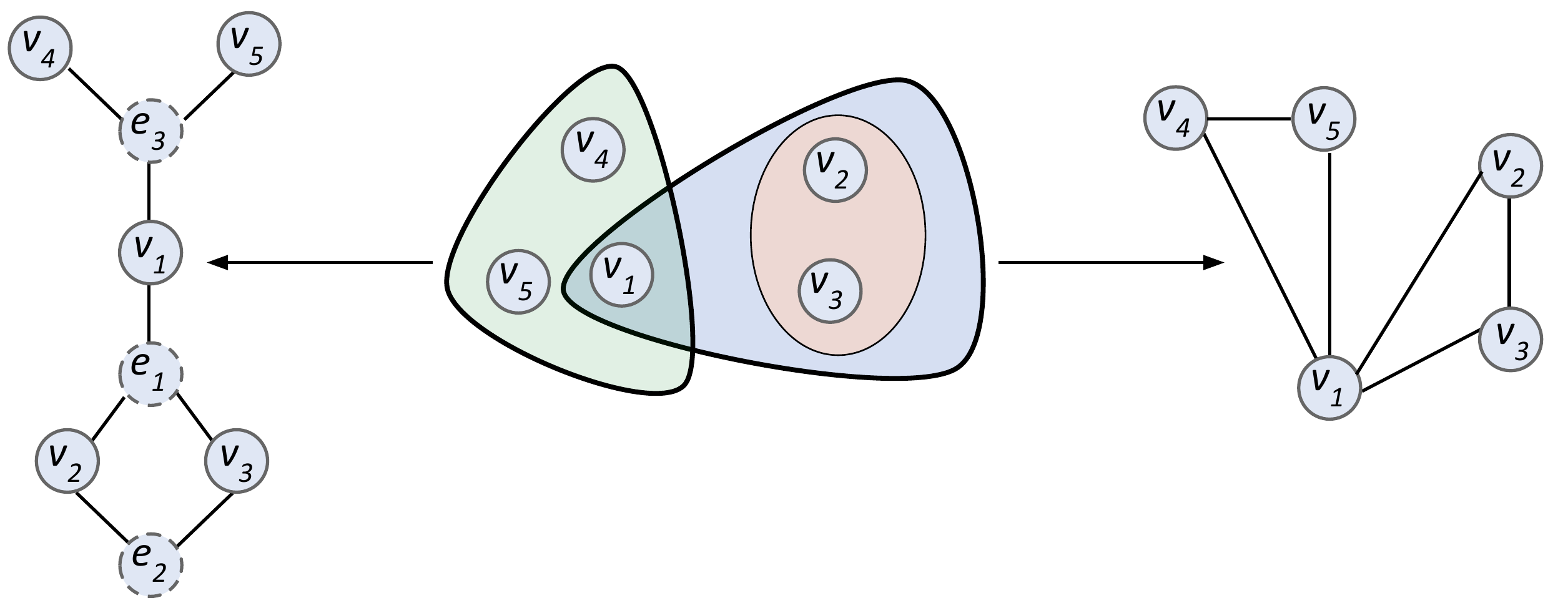}};
    \node[text width=3pt] (whitehead) at (1.2,0.32)
    {{\scriptsize clique}};
    \node[text width=3pt] (whitehead) at (1.1,0.0)
    {{\scriptsize expansion}};
    \node[text width=3pt] (whitehead) at (-2.3,0.32)
    {{\scriptsize star}};
    \node[text width=3pt] (whitehead) at (-2.5,0.0)
    {{\scriptsize expansion}};
    \node[text width=3pt] (whitehead) at (-0.7,-0.7)
    {{\scriptsize Hypergraph}};
    \end{tikzpicture}
    \vspace{-1em}
    \caption{}
    \label{fig_expansion}
\end{subfigure}\hfill
\begin{subfigure}{0.48\textwidth}
    \centering
    \includegraphics[scale=0.25]{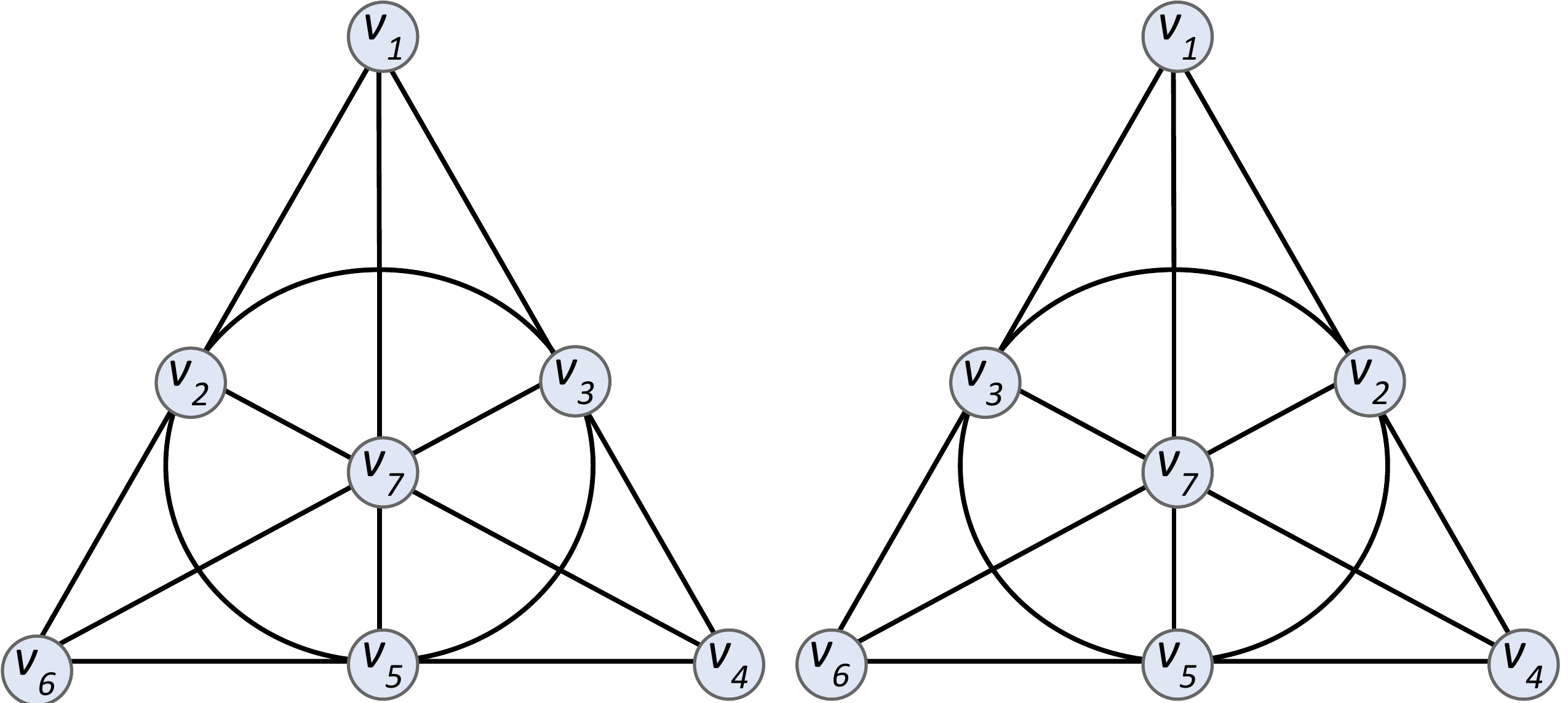}
    \caption{}
    \label{fig_fano}
\end{subfigure}
\caption{(a) Example showing reduction of a hypergraph to a graph using clique and star expansion methods. The clique expansion loses the unique information associated with the hyperedge defined by the set of nodes $\{v_2, v_3\}$, and it cannot distinguish it from the hyperedge defined by the nodes $\{v_1, v_2, v_3\}$. Star expansion  creates a heterogeneous graph that is difficult to handle using most well-studied graph methods \citep{hein2013total}. (b) Schematic representations of two Fano planes comprising 7 nodes and 7 hyperedges (6 straight lines and 1 circle.). The second Fano plane is a copy of the first with nodes $v_2$ and $v_3$ permuted. These two hypergraphs cannot be differentiated when transformed to a graph using clique expansion, thus making them indistinguishable.}
\label{fig1}
\end{figure}

Recent work in the field of geometric deep learning have presented formulations on graph structured data for the tasks of  node classification \citep{kipf2016semi}, link prediction \citep{zhang2018link}, or the classification of graphs \citep{zhang2018end}. Subsequently, for data containing higher-order relations, a few recent papers have presented hypergraph-based learning approaches on similar tasks \citep{yadati2019hypergcn, feng2019hypergraph}. A common implicit premise in these papers is that a hypergraph can be viewed as a specific type of regular graph. Therefore, reduction of hypergraph learning problem to that of a graph should suffice. Strategies to reduce a hypergraph to a graph include transforming the hyperedges into multiple edges using clique expansion \citep{feng2019hypergraph, jiang2019dynamic, zhang2018beyond}, converting  to a heterogeneous graph using star expansion \citep{agarwal2006higher}, and replacing every hyperedge with an edge created using a certain predefined metric \citep{yadati2019hypergcn}. Yet these methods are based on the wrong premise, motivated chiefly by a larger availability of graph-based approaches. By reducing a hypergraph to regular graph, these approaches make existing graph learning algorithms applicable to hypergraphs. However, hypergraphs are not a special case of regular graphs. The opposite is true, regular graphs are simply a specific type of hypergraph \citep{berge1976graphs}. Therefore, reducing the hypergraph problem to that of a graph cannot fully utilize the information available with hypergraph. Two schematic examples outlining this issue are shown in Fig.\ref{fig1}. To address tasks based on complex structured data, a hypergraph-based formulation is needed that complies with the properties of a hypergraph.

A major limitation of the existing hypergraph learning frameworks is their inherently transductive nature. This implies that these methods can only predict characteristics of nodes that were  present in the hypergraph at training time, and fail to infer on previously unseen nodes. The transductive nature of existing hypegraph approaches makes them inapplicable in, for example, finding the most promising target audience for a marketing campaign or making movie recommendations with new movies appearing all the time. An inductive solution would pave the way to solve such problems using hypergraphs. The inductive learning framework must be able to identify both the node’s local role in the hypergraph, as well as its global position \citep{hamilton2017inductive}. This is important for generalizing a newly observed hypergraph comprising previously unseen nodes to the learned node embeddings that the algorithm has optimized on, thus, making inductive learning a far more complex problem compared to the transductive learning methods.

In this paper, we address the above mentioned limitations of the existing hypergraph learning methods.
We propose a simple yet effective inductive learning framework for hypergraphs that is readily applicable to graphs as well. Our approach relies on neural message passing techniques due to which it can be used on hypergraphs of any degree of cardinality without the need for reduction to graphs. 
The points below highlight the contributions of this paper:
\begin{itemize}
    \item We address the challenging problem of representation learning on hypergraphs by proposing HyperSAGE, comprising a message passing scheme which is capable of jointly capturing the intra-relations (within a hyperedge) as well as inter-relations (across hyperedges). 
    \item The proposed hypergraph learning framework is inductive, i.e. it can perform predictions on previously unseen nodes, and can thus be used to model evolving hypergraphs.
    \item HyperSAGE facilitates neighborhood sampling and provides the flexibility in choosing different ways to aggregate information from the neighborhood.
    \item HyperSAGE is more stable than state-of-the-art methods, thus provides more accurate results on node classification tasks on hypergraphs with reduced variance in the output. 
\end{itemize}

\section{Related Work}
Learning node representations using graph neural networks has been a popular research topic in the field of geometric deep learning \citep{bronstein2017geometric}. Graph neural networks can be broadly classified into spatial (message passing) and spectral networks. We focus on a family of spatial message passing graph neural networks  that take a graph with some labeled nodes as input and learn embeddings for each node by aggregating information from its neighbors \citep{xu2018powerful}. Message passing operations in a graph simply propagate information along the edge connecting two nodes. Many variants of such message passing neural networks have been proposed, with some popular ones including \cite{gori2005new,li2015gated, kipf2016semi, gilmer2017neural,hamilton2017inductive}.

\cite{zhou2007learning} introduced learning on hypergraphs to model high-order relations for semi-supervised classification and clustering of nodes. Emulating a graph-based message passing framework for hypergraphs is not straightforward since a hyperedge involves more than two nodes which makes the interactions inside each hyperedge more complex. Representing a hypergraph with a matrix makes it rigid in describing the structures of higher order relations \citep{li2013z}. On the other hand, formulating message passing on a higher dimensional representation of hypergraph using tensors makes it computationally expensive and restricts it to only small datasets \citep{zhang2019introducing}. Several tensor based methods do perform learning on hypergraphs \citep{shashua2006multi,arya2019hyperlearn}, however they are limited to uniform hypergraphs only. 

To resolve the above issues,  \cite{feng2019hypergraph} and \cite{bai2020hypergraph} reduce a hypergraph to graph using clique expansion and perform graph convolutions on them. These  approaches cannot utilize complete structural information in the hypergraph and leads to unreliable learning performance for classifcation, clustering, active learning etc.\citep{li2017inhomogeneous,chien2019hs}. Another approach by \cite{yadati2019hypergcn}, named HyperGCN, replaces a hyperedge with pair-wise weighted edges between vertices (called mediators). With the use of mediators, HyperGCN can be interpreted as an improved approach of clique expansion, and to the best of our knowledge, is also the state-of-the-art method for hypergraph representation learning. However, for many cases such as Fano plane where each hyperedge contains at most three nodes, HyperGCN becomes equivalent to the clique expansion \citep{dong2020hnhn}. In spectral theory of hypergraphs, methods  have been proposed that fully exploit the hypergraph structure using non-linear Laplacian operators \citep{chan2018spectral,hein2013total}. In this work, we focus on message passing frameworks. Drawing inspiration from GraphSAGE \citep{hamilton2017inductive}, we propose to eliminate matrix (or tensor) based formulations in our neural message passing frameworks, which not only facilitates utilization of all the available information in a hypergraph, but also makes the entire framework inductive in nature.

\section{Proposed Model: HyperSAGE}
The core concept behind our approach is to aggregate feature information from the neighborhood of a node spanning across multiple hyperedges, where the edges can have varying cardinality. Below, we first define some preliminary terms, and then describe our generic aggregation framework. This framework performs message passing at two-levels for a hypergraph. Further, for any graph-structured data, our framework emulates the one-level aggregation similar to GraphSAGE \citep{hamilton2017inductive}.
Our approach inherently allows inductive learning, which makes it also applicable on hypergraphs with unseen nodes. 

\subsection{Preliminaries}
\textbf{Definition 1} (Hypergraph). \textit{A general hypergraph $\mathcal{H}$ can be represented as a pair $\mathcal{H}=(\mathcal{V, E})$, where $\mathcal{V}=\{v_1, v_2, ... , v_N\}$ denotes a set of $N$ nodes (vertices) and $\mathcal{E}=\{\mathbf{e}_1, \mathbf{e}_2, ... , \ \mathbf{e}_K\}$ denotes a set of hyperedges, with each hyperedge comprising a non-empty subset from $\mathcal{V}$. The maximum cardinality of the hyperedges in $\mathcal{H}$ is denoted as $M = \underset{\mathbf{e}\in\mathcal{E}}{\max}|\mathbf{e}|$.
}

Unlike in a graph, the hyperedges of $\mathcal{H}$ can contain different number of nodes and  $M$ denotes the largest number. From the definition above, we see that graphs are a special case of hypergraphs with $M$=2. Thus, compared to graphs, hypergraphs are  designed to model higher-order relations between nodes. Further, we define three types of neighborhoods in a hypergraph:

\textbf{Definition 2} (Intra-edge neighborhood). \textit{The intra-edge neighborhood of a node ${v_i} \in \mathcal{V}$ for any hyperedge $\mathbf{e} \in \mathcal{E}$ is defined as the set of nodes $v_j$ belonging to $\mathbf{e}$ and is denoted by  $\EuScript{N}(v_i, \mathbf{e})$  }

Further, let $E(v_i) = \{ \mathbf{e} \in \mathcal{E} \enskip | \enskip v_i \in \mathbf{e}\}$ be the sets of hyperedges that contain node $v_i$.

\textbf{Definition 3} (Inter-edge neighborhood). \textit{The inter-edge neighborhood of a node ${v_i} \in \mathcal{V}$ also referred as its global neighborhood, is defined as the neighborhood of ${v_i}$ spanning across the set of hyperedges $E(v_i)$ and is represented by $\EuScript{N}(v_i) = \bigcup_{\mathbf{e} \in E(v_i)}\EuScript{N}(v_i, \mathbf{e})$.}

\textbf{Definition 4} (Condensed neighborhood).\textit{ The condensed neighborhood of any node ${v_i} \in \mathcal{V}$ is a sampled set of $\alpha \leq |\mathbf{e}|$ nodes from a hyperedge $\mathbf{e} \in E(v_i)$ denoted by $N{(v_i,\mathbf{e};\alpha)}  \subset \mathcal{N}{(v_i,\mathbf{e})}$.}


\subsection{Generalized Message Passing Framework}



We propose to interpret the propagation of information in a given hypergraph as a two-level aggregation problem, where the neighborhood of any node is divided into \emph{intra-edge} neighbors and \emph{inter-edge} neighbors.
Given a hypergraph $\mathcal{H}=(\mathcal{V}, \mathcal{E})$, let $\mathbf{X}$ denote the feature matrix, such that $\mathbf{x}_i \in \mathbf{X}$ is the feature set for node $v_i \in \mathcal{V}$.  For two-level aggregation, let $\mathcal{F}_1(\cdot)$ and $\mathcal{F}_2(\cdot)$ denote the intra-edge and inter-edge aggregation functions, respectively. Message passing at node $v_i$ for aggregation of information at the $l^{\text{th}}$ layer can then be stated as

\begin{align}\label{eq:3}
    & \mathbf{x}_{i,l}^{(\mathbf{e})} \leftarrow \mathcal{F}_1\,(\{\mathbf{x}_{j,l-1}\: | \: v_j \in \mathcal{N}(v_i,\mathbf{e}; \alpha)\}), \\
    & \mathbf{x}_{i,l} \leftarrow \mathbf{x}_{i,l-1} + \mathcal{F}_2\,(\{\mathbf{x}_{i,l}^{(\mathbf{e})} \: | v_i \in E(v_i)\}),
\end{align}
where, $\mathbf{x}_{i,l}^{(\mathbf{e})}$ refers to the aggregated feature set at $v_i$ obtained with intra-edge aggregation for edge $\mathbf{e}$.
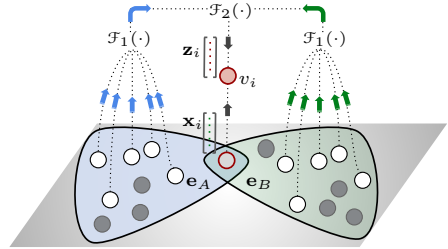
\begin{wrapfigure}{r}{0.5\textwidth}
\centering
\begin{tikzpicture}
    \node () at (0,0) {\includegraphics[scale=0.28]{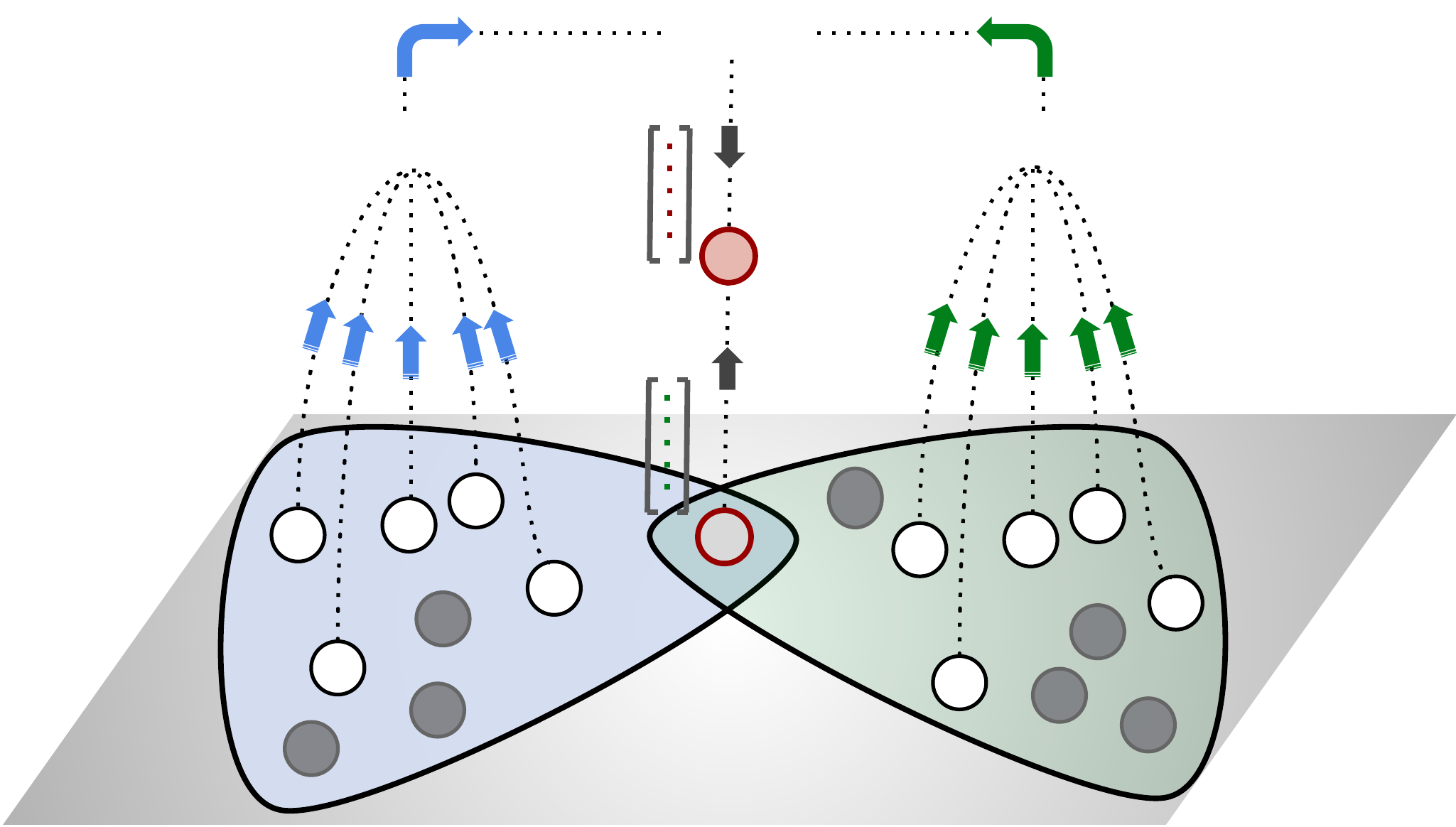}};
    \node[text width=3pt] (whitehead) at (1.05,1.12)
    {{\tiny $\mathcal{F}_1(\cdot)$}};
    \node[text width=3pt] (whitehead) at (-1.55,1.12)
    {{\tiny $\mathcal{F}_1(\cdot)$}};
    \node[text width=3pt] (whitehead) at (-0.2,1.49)
    {{\tiny $\mathcal{F}_2(\cdot)$}};
    \node[text width=3pt] (whitehead) at (-0.55,0.0)
    {{\scriptsize $\mathbf{x}_i$}};
    \node[text width=3pt] (whitehead) at (-0.55,0.95)
    {{\scriptsize $\mathbf{z}_i$}};
    \node[text width=3pt] (whitehead) at (0.2,0.55)
    {{\scriptsize $v_i$}};
    \node[text width=3pt] (whitehead) at (-0.5,-0.8)
    {{\scriptsize $\mathbf{e}_A$}};
    \node[text width=3pt] (whitehead) at (0.3,-0.8)
    {{\scriptsize $\mathbf{e}_B$}};
\end{tikzpicture}
\caption{Schematic representation of the two-level message passing scheme of HyperSAGE, with aggregation functions $\mathcal{F}_1(\cdot)$ and $\mathcal{F}_2(\cdot)$. It shows information aggregation from two hyperedges $\mathbf{e}_A$ and $\mathbf{e}_B$, where the intra-edge aggregation is from sampled sets of 5 nodes ($\alpha = 5$) for each hyperedge. }
\end{wrapfigure}

 The combined two-level message passing is then achieved using  nested aggregation function \mbox{$\mathcal{F} = \mathcal{F}_2(\mathcal{F}_1(\cdot))$}.
 To ensure that the expressive power of a hypergraph is preserved or at least the loss is minimized, the choice of aggregation function should comply with certain properties. 

Firstly, the aggregation function should be able to capture the features of neighborhood vertices in a manner that is invariant to the permutation of the nodes and hyperedges. Many graph representation learning methods use permutation invariant aggregation functions, such as $mean$, $sum$ and $max$ functions \citep{xu2018powerful}. These aggregations have proven to be successful for node classification problems. For the existing hypergraph frameworks, reduction to simple graphs along with a matrix-based message passing framework limits the possibilities of using different types of feature aggregation functions, and hence curtails the potential to explore unique node representations. 

Secondly, the aggregation function should also preserve the global neighborhood invariance at the `dominant nodes' of the graph. Here, dominant nodes refer to nodes that contain important features, thereby, impacting the learning process relatively more than their neighbors. The aggregation function should ideally be insensitive to the input, whether the provided hypergraph contains a few large hyperedges, or a larger number of smaller ones obtained from splitting them. Generally, a hyperedge would be split in a manner that the dominant nodes are shared across the resulting hyperedges. In such cases, global neighborhood invariance would imply that the aggregated output at these nodes before and after the splitting of any associated hyperedge stays the same. Otherwise, the learned representation of a node will change significantly with each hyperedge split.

Based on these considerations, we define the following properties for a generic message aggregation scheme that should hold for accurate propagation of information through the hypergraphs. 

\textbf{Property 1}  (Hypergraph Isomorphic Equivariance). \emph{A message aggregation scheme $\mathcal{F}$ would be equivariant to hypergraph isomporphism, if for two isomorphic hypergraphs $\mathcal{H}$ and \mbox{$\mathcal{H}^* = \sigma \bullet \mathcal{H}$}, the message aggregations can be expressed as $\mathcal{F}(\mathcal{H}^*) = \sigma \bullet \mathcal{F}(\mathcal{H})$, where $\sigma$ denotes a permutation operator on hypergraphs. }

\textbf{Property 2} (Global Neighborhood Invariance). \emph{The representation of a node after message passing should be invariant to the cardinality of the hyperedge, i.e., the aggregation scheme should not be sensitive to hyperedge contraction or expansion, as long as the global neighborhood of a node remains the same in the hypergraph.}

 The flexibility of our message passing framework allows us to go beyond the simple aggregation functions on hypergraphs without violating Property 1.  We introduce a series of power mean functions as aggregators, which have recently been shown to generalize  well on graphs \citep{li2020deepergcn}. We perform message aggregation in hypergraphs using these generalized means, denoted by $M_p$ and provide in section 4.2, a study on their performances. We also show that with appropriate combinations of the intra-edge and inter-edge aggregations Property 2 is also satisfied. 

\textbf{Aggregation Functions. }One major advantage of our strategy is that the message passing module is decoupled from the choice of the aggregation itself. This allows our approach to be used with a broad set of aggregation functions. We discuss below a few such possible choices.

\emph{Generalized means. }Also referred to as power means, this class of functions are very commonly used for getting an aggregated measure over a given set of samples. Mathematically, generalized means can be expressed as \mbox{$M_p = \left(\frac{1}{n}\sum_{i=1}^{n} x_i^p\right)^\frac{1}{p}$}, where $n$ refers to the number of samples in the aggregation, and $p$ denotes its power. The choice of $p$ allows providing different interpretations to the aggregation function. For example, $p=1$ denotes arithmetic mean aggregation, $p=2$ refers to mean squared estimate and a large value of $p$ corresponds to max pooling from the group. Similarly, $M_p$ can be used for geometric and harmonic means with $p \rightarrow 0$ and $p=-1$, respectively.

Similar to the recent work of \cite{li2020deepergcn}, we use generalized means for intra-edge as well as inter-edge aggregation. The two functions $\mathcal{F}_1(\cdot)$ and $\mathcal{F}_2(\cdot)$ for aggregation at node $v_i$ is defined as

\begin{align}
    \mathcal{F}_1(v_i) = \left(  \frac{1}{|\mathcal{N}(v_i,\mathbf{e})|} \sum_{v_j \in \mathcal{N}(v_i,\mathbf{e})}  \mathbf{x}_{j}^p\right)^{\frac{1}{p}},
    \mathcal{F}_2(v_i) = \left(  \frac{1}{|E(v_i)|} \sum_{\mathbf{e}\in E(v_i)}  \frac{\mathcal{N}(v_i,\mathbf{e})}{\mathcal{N}(v_i)}(\mathcal{F}_1({v_i}))^p\right)^{\frac{1}{p}}
    \label{eq_agg_mp}
\end{align}

 Note that in Eq. \ref{eq_agg_mp}, we have chosen the power term $p$ to be same for $\mathcal{F}_1$ and $\mathcal{F}_2$ so as to satisfy the global neighborhood invariance as stated in Property 2. Further, the scaling term $\frac{\mathcal{N}(v_i,\mathbf{e})}{\mathcal{N}(v_i)}$in $\mathcal{F}_2$ is added to balance the bias in the weighting introduced in intra-edge aggregation due to varying cardinality across the hyperedges.  These restrictions ensure that the joint aggregation $\mathcal{F}_2(\mathcal{F}_1(\cdot))$ satisfies the property of global neighborhood invariance at all times. Proof of the two aggregations satisfying Property 2 is stated in Appendix \ref{app_geom_agg}. 

\textbf{Sampling-based Aggregation.} Our neural message passing scheme provides the flexibility to adapt the message aggregation module to fit the desired computational budget through aggregating information from only a subset $N{(v_i,\mathbf{e};\alpha)}$ of the full neighborhood $N{(v_i,\mathbf{e})}$, if needed.  We propose to apply sub-sampling only on the nodes from the training set, and use information from the full neighborhood for the test set. The advantages of this are twofold. First, reduced number of samples per aggregation at training time reduces the relative computational burden. Second, similar to dropout \citep{srivastava2014dropout}, it serves to add regularization to the optimization process. Using the full neighborhood on test data avoids randomness in the test predictions, and generates consistent output.

\subsection{Inductive Learning on Hypergraphs}
HyperSAGE is a general framework for learning node representations on hypergraphs, on even unseen nodes. This approach uses a neural network comprising $L$ layers, and feature-aggregation is performed at each of these layers, as well as across the hyperedges. 
\begin{wrapfigure}{r}{0.5\textwidth}
\begin{minipage}[r]{0.5\textwidth}
\begin{algorithm}[H]
\SetAlgoLined
\SetKwInOut{Input}{Input}
\SetKwInOut{Output}{Output}
\Input{$\mathcal{H}=(\mathcal{V}, \mathcal{E})$; node features $\mathbf{X}$; depth $L$; weight matrices $\mathbf{W}^l$ for $l = 1 \hdots L$; non-linearity $\mathbf{\bm\sigma}$; intra-edge aggregation function $\mathcal{F}_1(\cdot)$; inter-edge aggregation function $\mathcal{F}_2(\cdot)$}
\Output{Node embeddings $\mathbf{z}_i | \enskip v_i \in \mathcal{V}$}
$\mathbf{h}_i^0 \leftarrow \mathbf{x}_i \in \mathbf{X} \enskip | \enskip v_i \in \mathcal{V}$\\
\For {$l = 1 \hdots L$}
{
\For {$\mathbf{e} \in \mathcal{E}$}{
    $\mathbf{h}_i^l \leftarrow \mathbf{h}_i^{l-1}$\\
    \For {$v_i \in \mathbf{e}$}
        {
            $\mathbf{h}_i^l \leftarrow \mathbf{h}_i^l + \mathcal{F}_2(\mathcal{F}_1(v_i))$ \\
        }  
    }
    $\mathbf{h}_i^l \leftarrow \sigma(\mathbf{W}^l(\mathbf{h}_i^l/||\mathbf{h}_i^l||_2)) \enskip | \enskip v_i \in \mathcal{V} $
 }
 $\mathbf{z}_i \leftarrow \mathbf{h}_i^L \enskip | \enskip v_i \in \mathcal{V} $
  \caption{HyperSAGE Message Passing}
  \label{algo1}
\end{algorithm}
\end{minipage}
\end{wrapfigure}
Algorithm \ref{algo1} describes the forward propagation mechanism which implements the aggregation function $\mathcal{F}(\cdot)=\mathcal{F}_2(\mathcal{F}_1(\cdot))$ described above. At each iteration, nodes first aggregate information from their neighbors within a specific hyperedge. This is repeated over all the hyperedges across all the $L$ layers of the network. The trainable weight matrices $\mathbf{W}^l$ with $l \in L$ are used to aggregate information across  the feature dimension and propagate it through the various layers of the hypergraph.

\textbf{Generalizability of HyperSAGE.} HyperSAGE can be interpreted as a generalized formulation that unifies various existing graph-based as well as hypergraph formulations. Our approach unifies them, identifying each of these as special variants/cases of our method. We discuss here briefly the two popular algorithms.

\textit{Graph Convolution Networks (GCN).} The GCN approach proposed by \cite{kipf2016semi} is a graph-based method that can be derived as a special case of HyperSAGE with maximum cardinality $|M|=2$, and setting the agggregation function $\mathcal{F}_2 = M_p$ with $p=1$. As this being a graph-based method, $\mathcal{F}_1$ will not be used.

\textit{GraphSAGE.} Our approach, when reduced for graphs using $|M|=2$, is similar to GraphSAGE. For exact match, the aggregation function $\mathcal{F}_2$ should be one of $mean$, $max$ or $LSTM$. Further, the sampling term $\alpha$ can be adjusted to match the number of samples per aggregation as in GraphSAGE.

\section{Experiments}


\subsection{Experimental Setup}

For the experiments in this paper, we use co-citation and co-authorship network datasets: CiteSeer, PubMed, Cora \citep{sen2008collective} and DBLP \citep{nr}. The task for each dataset is to predict the topic to which a document belongs (multi-class classification). Further, for all experiments, we use a neural network with 2 layers. All models are implemented in Pytorch and trained using Adam optimizer. Additional implementation details are presented in Appendix \ref{app_sec_imp_details}.

\subsection{Semi-supervised Node Classification on Hypergraphs}

\textbf{Performance comparison with existing methods. }We implemented HyperSAGE for the task of semi-supervised classification of nodes on a hypergraph, and the results are compared with state-of-the art methods. These include (a) Multi-layer perceptron with explicit hypergraph Laplacian regularisation (MLP + HLR), (b) Hypergraph Neural Networks (HGNN) \citep{feng2019hypergraph} which uses a clique expansion, and (c) HyperGCN and its variants \citep{yadati2019hypergcn} that collapse the hyperedges using mediators. For HyperSAGE method, we use 4 variants of generalized means $M_p$ with $p = 1, 2, -1$ and $0.01$ with complete neighborhood i.e., $\alpha = |\mathbf{e}|$.  For all the cases, 10 data splits over 8 random weight initializations are used, totalling 80 experiments per method and for every dataset. The data splits are same as in HyperGCN described in Appendix \ref{app_sec_dataset}.

\begin{table}
\centering
\caption{Performance of HyperSAGE and other hypergraph learning methods on co-authorship and co-citation datasets. }
\begin{tabular}{lccccc}
\toprule
  &\multicolumn{2}{c}{\textbf{Co-authorship Data}} & \multicolumn{3}{c}{\textbf{Co-citation Data}} \\ 
\cmidrule(lr){2-3} \cmidrule(lr){4-6} 
  \textbf{Method} &DBLP &Cora&Pubmed &Citeseer & Cora \\
\cmidrule(l){1-1}\cmidrule(l){2-2} \cmidrule(l){3-3} \cmidrule(l){4-4} \cmidrule(l){5-5} \cmidrule(l){6-6}
MLP $+$ HLR               &   63.6 $\pm$ 4.7             &    59.8 $\pm$ 4.7             &     64.7 $\pm$ 3.1             &     56.1 $\pm$ 2.6    &      61.0 $\pm$ 4.1\\ [3pt]
HGNN                    & 69.2 $\pm$ 5.1                &    63.2 $\pm$ 3.1             & 66.8 $\pm$ 3.7                 & 56.7 $\pm$ 3.8
& \textbf{70.0 $\pm$ 2.9}                \\ [3pt]
FastHyperGCN            & 68.1 $\pm$ 9.6                & 61.1 $\pm$ 8.2               & 65.7 $\pm$ 11.1                  & 56.2 $\pm$ 8.1
& 61.3 $\pm$ 10.3                \\ [3pt]
HyperGCN                & 70.9 $\pm$ 8.3               & 63.9 $\pm$ 7.3               & 68.3 $\pm$ 9.5      & 57.3 $\pm$ 7.3          & 62.5 $\pm$ 9.7                \\ [3pt]
\midrule
HyperSAGE ($p=2$)              & 71.5 $\pm$ 4.4              & 69.8 $\pm$ 2.6                &  71.3 $\pm$ 2.4      &       59.8 $\pm$ 3.3 &62.9 $\pm$ 2.1               \\ 
HyperSAGE ($p=1$)              & 77.2 $\pm$ 4.3               & \textbf{72.4 $\pm$ 1.6}                &  72.6 $\pm$ 2.1      &       \textbf{61.8 $\pm$ 2.3} & 69.3 $\pm$ 2.7               \\
HyperSAGE ($p=0.01$)              & \textbf{77.4 $\pm$ 3.8}               & 72.1 $\pm$ 1.8               &  \textbf{72.9 $\pm$ 1.3}       &       61.3 $\pm$ 2.4 & 68.2 $\pm$ 2.4               \\
HyperSAGE ($p=-1$)              & 70.9 $\pm$ 2.3              & 67.4 $\pm$ 2.1                &  68.3 $\pm$ 3.1      &       59.8 $\pm$ 2.0 & 62.3 $\pm$ 5.7               \\

 \bottomrule
\end{tabular}
\label{table_exp1}
\end{table}

Table \ref{table_exp1} shows the results obtained for the node classification task. We see that the different variants of HyperSAGE consistently show better scores across our benchmark datasets, except Cora co-citation where no improvement is observed compared to HGNN. Cora co-citation data is relatively very small in size with a cardinality of $3.0 \pm 1.1$, and we speculate that there does not exist enough scope of improving with HyperSAGE beyond what HGNN can express with the clique expansion.

For the larger datasets such as DBLP and Pubmed, we see that the improvements obtained in performance with HyperSAGE over the best baselines are 6.3\% and 4.3\% respectively. Apart from its superior performance, HyperSAGE is also stable, and is less sensitive to the choice of data split and initialization of the weights. This is evident from the scores of standard deviation (SD) for the various experiments in Table \ref{table_exp1}. We see that the SD scores for our method are lower than other methods, and there is a significant gain in performance compared to HyperGCN. Another observation is that the HyperGCN method is very sensitive to the data splits as well as initializations with very large errors in the predictions. This is even higher for the FastHyperGCN variant. Also, we have found that all the 4 choices of $p$ work well with HyperSAGE for these datasets. We further perform a more comprehensive study analyzing the effect of $p$ on model performance in the next experiment.

\begin{figure}%
    \centering
    \begin{subfigure}{0.24\textwidth}
    \includegraphics[width=3.3cm]{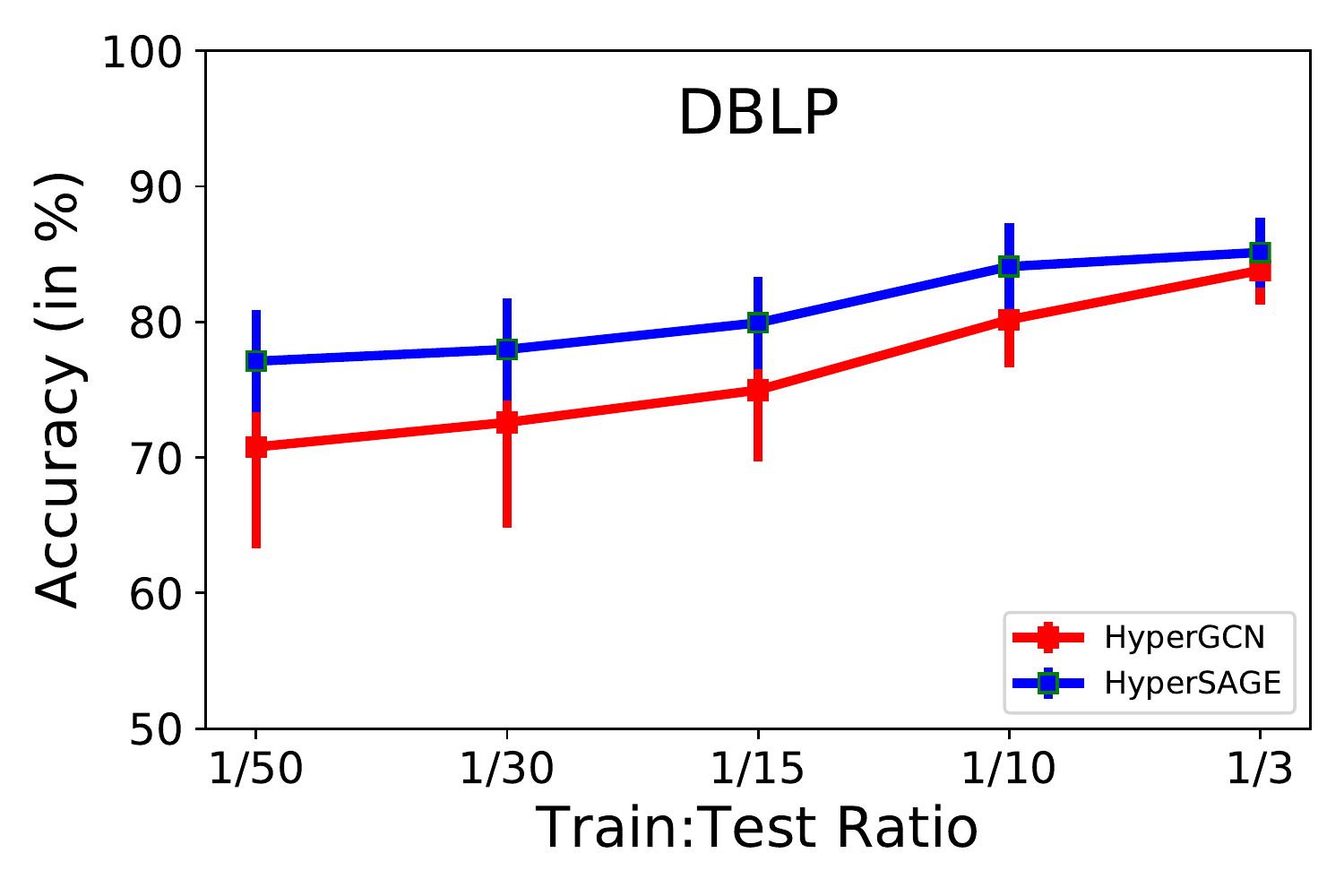}%
    \end{subfigure}
    \begin{subfigure}{0.24\textwidth}
    \includegraphics[width=3.3cm]{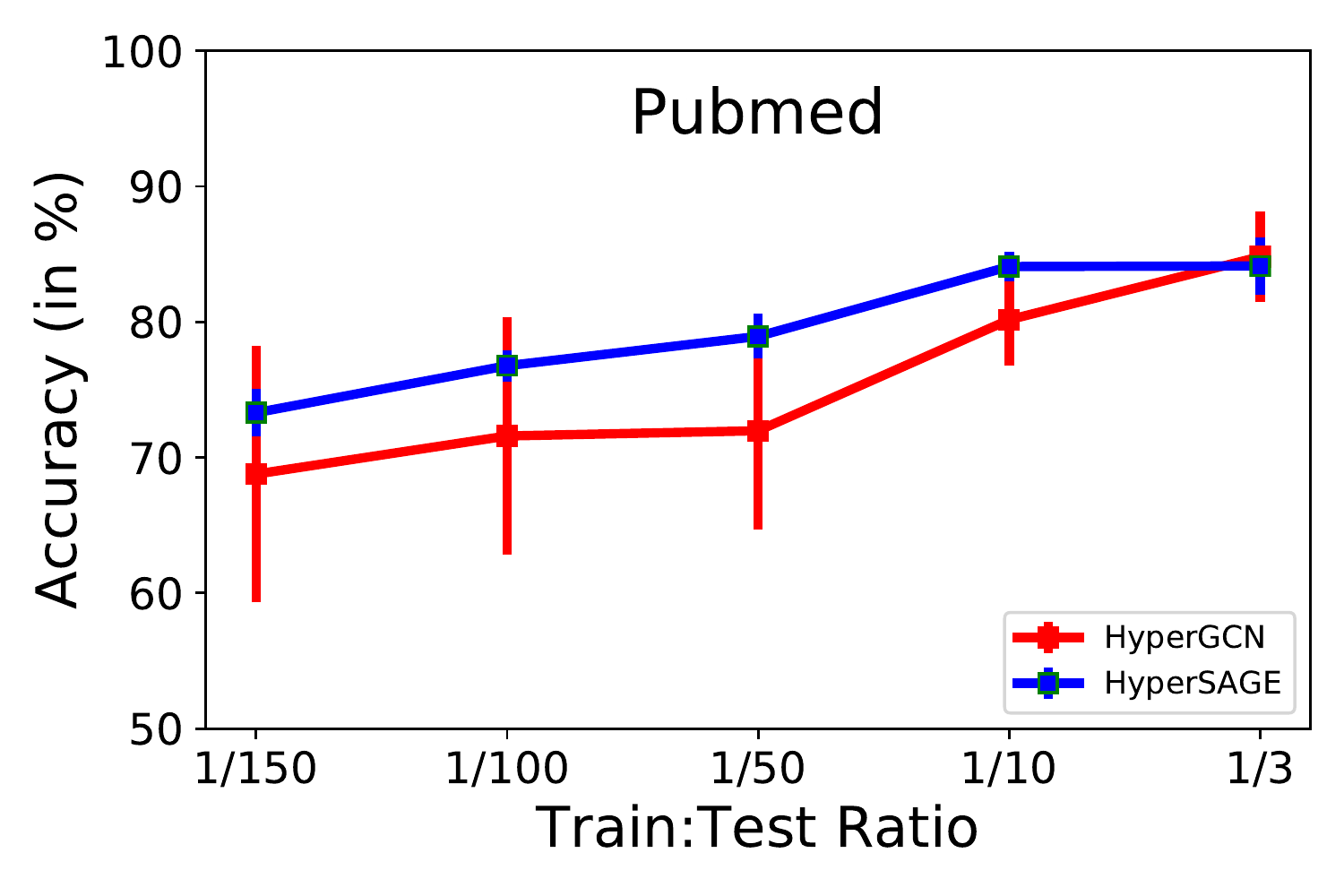}%
    \end{subfigure}
    \begin{subfigure}{0.24\textwidth}
    \includegraphics[width=3.3cm]{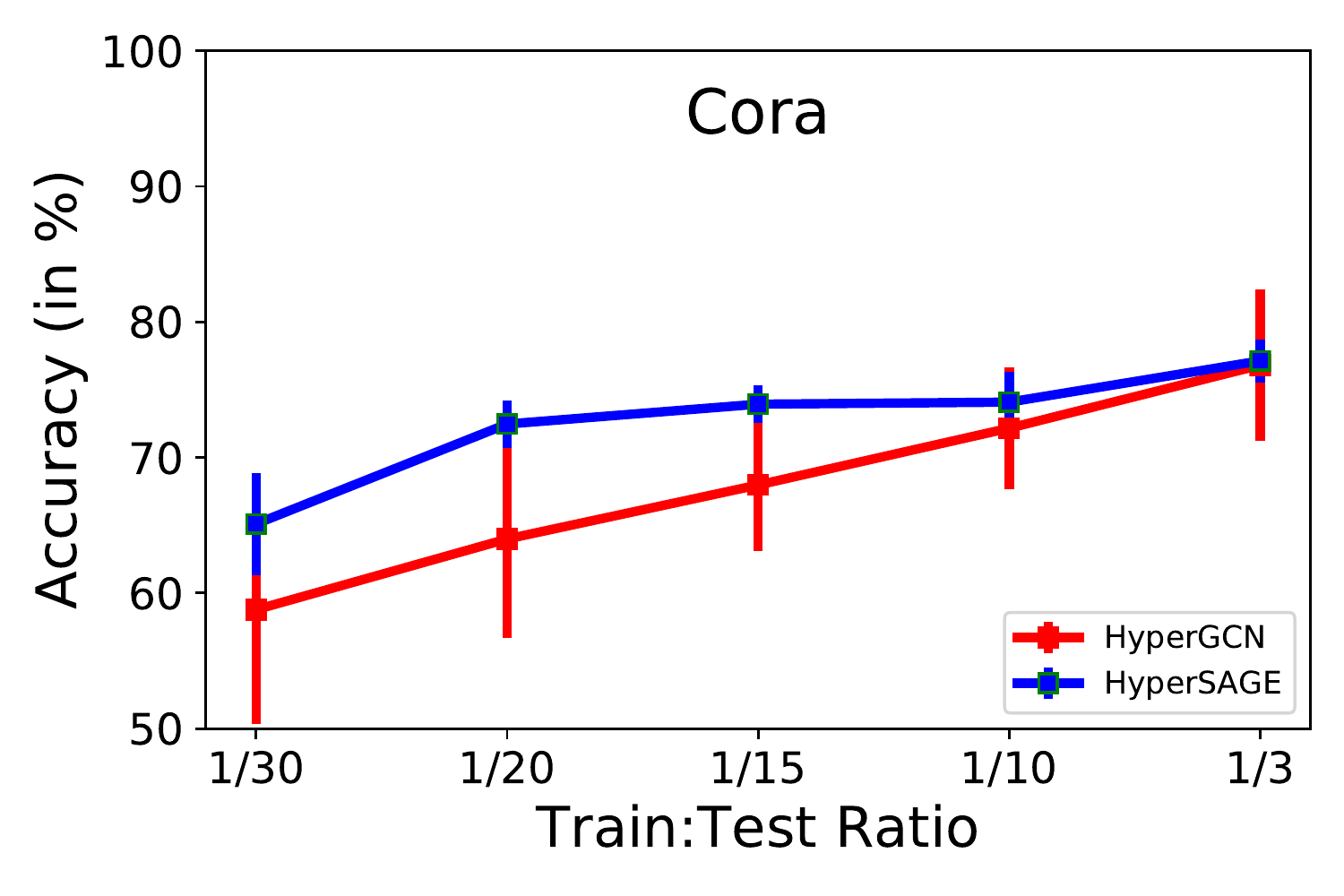}%
    \end{subfigure}
    \begin{subfigure}{0.24\textwidth}
    \includegraphics[width=3.3cm]{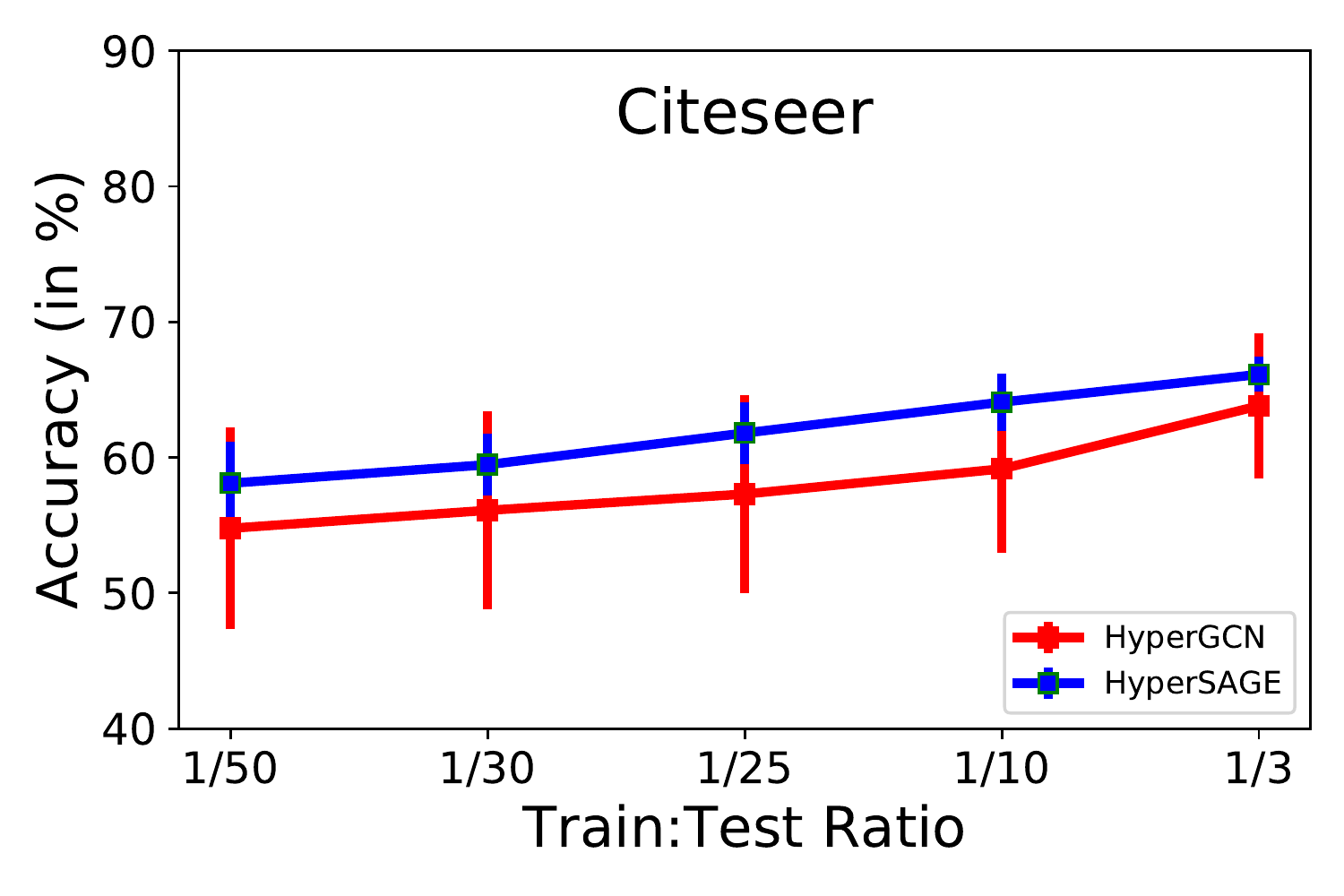}%
    \end{subfigure}
     \caption{Accuracy scores for HyperSAGE and HyperGCN obtained for different train-test ratios for multi-class classification datasets.}
    \label{fig_stable}%
\end{figure}

\textbf{Stability analysis. }We further study the stability of our method in terms of the variance observed in performance for different ratios of train and test splits, and compare results with that of HyperGCN implemented under similar settings. Fig. \ref{fig_stable} shows results for the two learning methods on 5 different train-test ratios. We see that the performance of both models improves when a higher fraction of data is used for training, and the performances are approximately the same at the train-test ratio of 1/3. However, for smaller ratios, we see that HyperSAGE outperforms HyperGCN by a significant margin across all datasets. Further, the standard deviation for the predictions of HyperSAGE are significantly lower than that of HyperGCN. Clearly, this implies that HyperSAGE is able to better exploit the information contained in the hypergraph compared to HyperGCN, and can thus produce more accurate and stable predictions. Results on Cora and Citeseer can be found in Appendix \ref{app:result}.

\begin{table}[t]
\centering
\caption{Performance of HyperSAGE for multiple values of $p$ in generalized means aggregator ($M_p$) on varying number of neighborhood samples ($\alpha$).}
\begin{tabular}{lcccccccc}
\toprule
  &\multicolumn{4}{c}{DBLP} & \multicolumn{4}{c}{Pubmed} \\ 
\cmidrule(lr){2-5} \cmidrule(lr){6-9} 
  &$\alpha = 2$ &$\alpha = 3$ &$\alpha = 5$ &$\alpha = 10$ &$\alpha = 2$ &$\alpha = 3$ &$\alpha = 5$ &$\alpha = 10$\\
 \cmidrule(l){2-2} \cmidrule(l){3-3} \cmidrule(l){4-4} \cmidrule(l){5-5}  \cmidrule(l){6-6} \cmidrule(l){7-7} \cmidrule(l){8-8} \cmidrule(l){9-9}
 
$p = -1$ & 59.6& 61.2 &69.9 &70.9 &60.1 & 60.2 & 67.9 & 66.4  \\
$p = 0.01$ & 61.2 & 64.8 & 73.1 & \textbf{77.4} & 65.5 & 67.4   & \textbf{73.4}& 72.9 \\
$p = 1$ & 62.3 & 64.5 & 73.1 &77.2 &64.8 & 64.3 &72.2 & 72.6 \\
$p = 2$ & 63.1 & 63.8 & 71.9 & 71.5 & 63.7 & 63.9 & 70.8& 71.3\\
$p = 3$ & 62.7 & 63.6 &71.3 & 71.4 & 62.2 & 61.3 & 70.1 & 67.9 \\
$p = 5$ & 62.8&63.3 &69.4 & 70.6 &62.1 &60.4 &69.3 & 68.0\\
 
\bottomrule
\end{tabular}

\label{tab_alphap}

\end{table}

\textbf{Effect of generalized mean aggregations and neighborhood sampling.} We study here the effect of different choices of the aggregation functions $\mathcal{F}_1(\cdot)$ and $\mathcal{F}_2(\cdot)$ on the performance of the model. Further, we also analyze how the number of samples chosen for aggregation affect its performance. Aggregation functions from $M_p$ are chosen with $p=1,2,3,4,5, 0.01$ and $-1$, and to comply with global neighborhood invariance, we use the same value of $p$ for both aggregation functions as stated in Eq. \ref{eq_agg_mp}. The number of neighbors $\alpha$ for intra-edge aggregation are chosen to be 2, 3, 5 and 10. Table \ref{tab_alphap} shows the accuracy scores obtained for different choices of $p$ and $\alpha$ on DBLP and Pubmed datasets. For most cases, higher value of $p$ reduces the performance of the model. For $\alpha=2$ on DBLP, performance seems to be independent of the choice of $p$. A possible explanation could be that the number of neighbors is very small, and change in $p$ does not affect the propagation of information significantly. An exception is $p=-1$, where the performance drops for all cases. For Pubmed, the choice of $p$ seems to be very important, and we find that $p=0.01$ seems to fit best. 

We also see that the number of samples per aggregation can significantly affect the performance of the model. For DBLP, model performance increases with increasing value of $\alpha$. However, for Pubmed, we observe that performance improves up to $\alpha=5$, but then a slight drop is observed for larger sets of neighbors. Note that for Pubmed, the majority of the hyperedges have cardinality less than or equal to 10. This means that during aggregation, information will most often be aggregated from all the neighbors, thereby involving almost no stochastic sampling. Stochastic sampling of nodes could serve as a regularization mechanism and reduce the impact of noisy hyperedges. However, at $\alpha=10$, it is almost absent, due to which the noise in the data affects the performance of the model which is not the case in DBLP.


\begin{table}[t]
\centering
\caption{Performance of HyperSAGE and its variants on nodes which were part of the training hypergraph (seen) and nodes which were not part of the training hypergraph (unseen).}
\begin{tabular}{lcccccccc}
\toprule
  &\multicolumn{2}{c}{DBLP} & \multicolumn{2}{c}{Pubmed} &\multicolumn{2}{c}{Citeseer} & \multicolumn{2}{c}{Cora (citation)}\\ 
 \cmidrule(lr){2-3} \cmidrule(lr){4-5} \cmidrule(lr){6-7} \cmidrule(lr){8-9}
  \textbf{Method} &Seen &Unseen &Seen &Unseen &Seen &Unseen &Seen &Unseen\\
 \cmidrule(l){2-2} \cmidrule(l){3-3} \cmidrule(l){4-4} \cmidrule(l){5-5}  \cmidrule(l){6-6} \cmidrule(l){7-7} \cmidrule(l){8-8} \cmidrule(l){9-9}
 
MLP + HLR & 64.5 &58.7 &66.8 &62.4 &60.1 &58.2 &65.7 &64.2  \\
HyperSAGE $(p = 0.01)$ & 78.1 & 73.1 & 81.0 &80.4 & 69.2 & 67.1 &68.2 &65.7  \\
HyperSAGE $(p = 1)$ & 78.1 & 73.2 &78.5 &76.4  & 69.3&67.9 & 71.3&66.8 \\
HyperSAGE $(p = 2)$ & 76.1 &70.2 &71.2 &69.8  & 65.9 & 63.8  &65.9 &64.5 \\
 
\bottomrule
\end{tabular}

\label{tab_inductive}

\end{table}

\subsection{Inductive learning on evolving graphs}

For inductive learning experiment, we consider the case of evolving hypergraphs. We create 4 inductive learning datasets from DBLP, Pubmed, Citeseer and Core (co-citation) by splitting each of the datasets into a train-test ratio of 1:4. Further, the test data is split into two halves: \emph{seen} and \emph{unseen}. The seen test set comprises nodes that are part of the hypergraph used for representation learning. Further, unseen nodes refer to those that are never a part of the hypergraph during training. To study how well HyperSAGE generalizes for inductive learning, we classify the unseen nodes and compare the performance with the scores obtained on the seen nodes. Further, we also compare our results on unseen nodes with those of MLP+HLR. The results are shown in Table \ref{tab_inductive}. We see that results obtained with HyperSAGE on unseen nodes are significantly better than the baseline method. Further, these results seem to not differ drastically from those obtained on the seen nodes, thereby confirming that HyperSAGE can work with evolving graphs as well.  


\section{Conclusion}
We have proposed HyperSAGE, a generic neural message passing framework for inductive learning on hypergraphs. The proposed approach fully utilizes the inherent higher-order relations in a hypergraph structure without reducing it to a regular graph. Through numerical experiments on several representative datasets, we show that HyperSAGE outperforms the other methods for hypergraph learning. Several variants of graph-based learning algorithm such as GCN and GraphSAGE can be derived from the flexible aggregation and neighborhood sampling framework, thus making HyperSAGE a universal framework for learning node representation on hypergraphs as well as graphs.

\subsubsection*{Acknowledgments}
This research has received funding from the European Union’s
Horizon 2020 research and innovation programme under grant
agreement 700381.

\bibliography{iclr2021_conference}
\bibliographystyle{iclr2021_conference}
\section*{Appendices}
\appendix

\section{Experiments: Additional details}
 We perform multi-class classification on co-authorship and co-citation datasets, where the task is to predict the topic (class) for each document.
\subsection{Dataset Description}
Hypergraphs are created on these datasets by assigning each document as a node and each hyperedge represents (a) all documents co-authored by an author in co-authorship dataset and (b) all documents cited together by a document in co-citation dataset. Each document (node) is represented by bag-of-words features. The details about nodes, hyperedges and features is shown in Table \ref{dataset}.  We use the same dataset and train-test splits as provided by \cite{yadati2019hypergcn} in their publically available implementation \footnote{HyperGCN Implementation: https://github.com/malllabiisc/HyperGCN}.

\label{app_sec_dataset}
\begin{table}[]
\centering
\caption{Details of real-world hypergraph datasets used in our work}
\begin{tabular}{lccccc}
\toprule
  &\multicolumn{2}{c}{\textbf{Co-authorship Data}} & \multicolumn{3}{c}{\textbf{Co-citation Data}} \\ 
\cmidrule(lr){2-3} \cmidrule(lr){4-6} 
  &DBLP &Cora&Pubmed &Citeseer & Cora \\
\cmidrule(l){2-2} \cmidrule(l){3-3} \cmidrule(l){4-4} \cmidrule(l){5-5} \cmidrule(l){6-6}
Nodes ($|\mathcal{V}|$) & 43413 & 2708 & 19717 & 3312 & 2708 \\ [3pt]
Hyperedges ($|\mathcal{E}|$) & 22535 & 1072 & 7963 & 1079 & 1579\\[3pt]
average hyperedge size &4.7$\pm$6.1 &4.2$\pm$4.1  & 4.3 $\pm$ 5.7 &3.2$\pm$2.0 & 3.0 $\pm$ 1.1  \\ [3pt]
number of features, $|\mathbf{x}|$ &1425&1433&500&3703&1433 \\ [3pt]
number of classes &6&7&3&6&7\\
\bottomrule
\end{tabular}
\label{dataset}
\end{table}

\subsection{Implementation details}
\label{app_sec_imp_details}
We use the following set of hyperparameters similar to the prior work by \cite{kipf2016semi} for all the models.
\begin{itemize}
    \item hidden layer size: 32
    \item dropout rate: 0.5
    \item learning rate: 0.01
    \item weight decay: 0.0005
    \item number of training epochs: 150
    \item $\lambda$ for explicit Laplacian regularisation: 0.001
    
\end{itemize}

\section{Choice of inter-edge and intra-edge aggregations}
\label{app_geom_agg}

\emph{Proof. } For any given hypergraph $\mathcal{H}_1=(\mathcal{V}, \mathcal{E}_1)$, we consider that a node $v_i$ exists, such that the number of hyperedges containing $v_i$ is given by $|E(v_i)|$. The aggregation output $\mathcal{F}_1(v_i)$ can then be written using generalized means $M_p$ as  

\begin{equation}
    \mathcal{F}_1(v_i) = \left(  \frac{1}{|\mathcal{N}(v_i,\mathbf{e})|} \sum_{v_j \in \mathcal{N}(v_i,\mathbf{e})}  \mathbf{x}_{j}^{p_1}\right)^{\frac{1}{p_1}}.
    \label{intra_eq1}
\end{equation}
Further, the inter-edge aggregation $\mathcal{F}_2(\cdot)$ can be stated as

\begin{equation}
        \mathcal{F}_2(v_i) = \left(  \frac{1}{|E(v_i)|} \sum_{\mathbf{e}\in E(v_i)} 
       \left(   \frac{1}{|\mathcal{N}(v_i,\mathbf{e})|} \sum_{v_j \in \mathcal{N}(v_i,\mathbf{e})}  \mathbf{x}_{j}^{p_1}\right)^{\frac{p_2}{p_1}}    \right)^{\frac{1}{p_2}}
\end{equation}

In order to prove the global neighborhood equivariance property, the above equation can be rewritten as 

\begin{equation}\label{expanded}
        \mathcal{F}_2(v_i) = \left(  \frac{1}{|E(v_i)|} \left(
        \left(   \frac{1}{|\mathcal{N}(v_i,\mathbf{e}_q)|} \sum_{v_j \in \mathcal{N}(v_i,\mathbf{e}_q)}  \mathbf{x}_{j}^{p_1}\right)^{\frac{p_2}{p_1}} + 
        \sum_{\mathbf{e}\in E(v_i),\mathbf{e}\neq \mathbf{e}_q} 
       \left(   \frac{1}{|\mathcal{N}(v_i,\mathbf{e})|} \sum_{v_j \in \mathcal{N}(v_i,\mathbf{e})}  \mathbf{x}_{j}^{p_1}\right)^{\frac{p_2}{p_1}}\right)    \right)^{\frac{1}{p_2}}
\end{equation}
Further, let 

\begin{equation}
    \Psi = \sum_{\mathbf{e}\in E(v_i),\mathbf{e}\neq \mathbf{e}_q} 
       \left(   \frac{1}{|\mathcal{N}(v_i,\mathbf{e})|} \sum_{v_j \in \mathcal{N}(v_i,\mathbf{e})}  \mathbf{x}_{j}^{p_1}\right)^{\frac{p_2}{p_1}},
\end{equation}
then Eq. \ref{expanded} can be rewritten as

\begin{equation}
        \mathcal{F}_2(v_i) = \left(  \frac{1}{|E(v_i)|} \left(
        \left(   \frac{1}{|\mathcal{N}(v_i,\mathbf{e}_q)|} \sum_{v_j \in \mathcal{N}(v_i,\mathbf{e}_q)}  \mathbf{x}_{j}^{p_1}\right)^{\frac{p_2}{p_1}} + 
        \Psi\right)    \right)^{\frac{1}{p_2}}
\end{equation}

Let us assume now that hyperedge $\mathbf{e}_q$ is split into $r$ hyperedges given by  $E(v_i,\mathbf{e}_q)=\{\mathbf{e}_{q_1}, \mathbf{e}_{q_2} \hdots \mathbf{e}_{q_r}$\}. Thus, splitting the local neighborhood of $v_i$ with the global neighborhood still the same. Stating the aggregation on the new set of hyperedges as $\Tilde{\mathcal{F}}_2(v_i)$, we assemble the contribution from this new set of hyperedges with added weight terms as stated below.

\begin{equation}
        \Tilde{\mathcal{F}}_2(v_i) = \left(  \frac{1}{|E(v_i)|} \left(
       \sum_{\mathbf{e}\in E(v_i,\mathbf{e}_q)} 
       \left(   \frac{w_q}{|\mathcal{N}(v_i,\mathbf{e})|} \sum_{v_j \in \mathcal{N}(v_i,\mathbf{e})}  \mathbf{x}_{j}^{p_1}\right)^{\frac{p_2}{p_1}} + 
        \Psi\right)    \right)^{\frac{1}{p_2}}
\end{equation}

For global neighborhood invariance, we should satisfy the condition  $\mathcal{F}_2(v_i) = \Tilde{\mathcal{F}}_2(v_i)$. Based on this, we would like to solve for the weights $w_q$.
\begin{equation}
  \left(   \frac{1}{|\mathcal{N}(v_i,\mathbf{e}_q)|} \sum_{v_j \in \mathcal{N}(v_i,\mathbf{e}_q)}  \mathbf{x}_{j}^{p_1}\right)^{\frac{p_2}{p_1}} = \sum_{\mathbf{e}\in E(v_i,\mathbf{e}_q)} 
       \left(   \frac{w_q}{|\mathcal{N}(v_i,\mathbf{e})|} \sum_{v_j \in \mathcal{N}(v_i,\mathbf{e})}  \mathbf{x}_{j}^{p_1}\right)^{\frac{p_2}{p_1}}
\end{equation}

To obtain a feasible solution for the above equation, we first assume $p_1 = p_2$. Further, we match the terms  and then solve for $w_q$ from the following equation.
\begin{equation}
\frac{1}{|\mathcal{N}(v_i,\mathbf{e}_q)|} \sum_{v_j \in \mathcal{N}(v_i,\mathbf{e}_q)}  \mathbf{x}_{j}^{p} = 
\sum_{\mathbf{e}\in E(v_i,\mathbf{e}_q)} 
\frac{w_q}{|\mathcal{N}(v_i,\mathbf{e})|} \sum_{v_j \in \mathcal{N}(v_i,\mathbf{e})}  \mathbf{x}_{j}^{p}
\end{equation}

We assume here that every other neighbor of $v_i$ in hyperedge $\mathbf{e}_q$ is only assigned to one of the hyperedges in $E(v_i, \mathbf{e}_q)$. Based on this, we can see that equate the coefficients of $\mathbf{x}_j^p$ to solve for $w_q$. Thus, we have
\begin{equation}
\frac{1}{|\mathcal{N}(v_i,\mathbf{e}_q)|} = 
\frac{w_q}{|\mathcal{N}(v_i,\mathbf{e})|}.
\end{equation}

Rearranging the terms, we obtain

\begin{equation}
w_q = \frac{|\mathcal{N}(v_i,\mathbf{e})|}{|\mathcal{N}(v_i,\mathbf{e}_q)|} \enskip \forall \enskip \mathbf{e} \in E(v_i, \mathbf{e}_q).
\end{equation}

Thus, if an edge $\mathbf{e}_q$ is split into multiple edges $E(v_i, \mathbf{e}_q)$, then for the two aggregations to hold, the conditions are $p_1 = p_2$ and $w_q = \frac{|\mathcal{N}(v_i,\mathbf{e})|}{|\mathcal{N}(v_i,\mathbf{e}_q)|} \enskip \forall \enskip \mathbf{e} \in E(v_i, \mathbf{e}_q)$.

Next, interpreting that $E(v_i)$ is a set of hyperedges that are obtained from a single hyperedge, the condition can similarly be translated as $p_1 = p_2$ and $w_q = \frac{|\mathcal{N}(v_i,\mathbf{e})|}{|\mathcal{N}(v_i)|} \enskip \forall \enskip \mathbf{e} \in E(v_i)$. The notation for this can be seen in Eq. \ref{eq_agg_mp} in the main text.



\end{document}